# Lattice-Based Graded Logic: A Multimodal Approach

**Philippe Chatalic** **Christine Froidevaux**

Laboratoire de Recherche en Informatique - UA CNRS 410
Bat 490 - Université Paris-Sud - 91405 ORSAY Cedex - FRANCE
{chatalic, chris}@lri.lri.fr

## Abstract

Experts do not always feel very comfortable when they have to give precise numerical estimations of certainty degrees. In this paper we present a qualitative approach which allows for attaching partially ordered symbolic grades to logical formulas. Uncertain information is expressed by means of parameterized modal operators. We propose a semantics for this multimodal logic and give a sound and complete axiomatization. We study the links with related approaches and suggest how this framework might be used to manage both uncertain and incomplete knowledge.

## 1. INTRODUCTION

Intelligent decision support systems often have to deal with uncertain knowledge. In most of them, numerical degrees are used for quantifying uncertainty. Various mathematical settings have been proposed for representing levels of certainty, e.g., Probability theory, Dempster-Shafer theory [Shafer 1976] and Possibility theory [Zadeh 1978]. However, experts do not always feel very comfortable when they have to give precise estimations of certainty degrees. Unless it is possible to rely on statistical studies, criteria involved in such estimations remain very peculiar to each expert. This is perhaps the reason why experts are more readily willing to provide qualitative estimations. In such cases, instead of giving precise estimates, they are just required to express when a statement is more certain than another. Several attempts have proposed qualitative views of numerical settings (e.g., [Fine 1973], [Gärdenfors 1975], [Halpern & Rabin 1987], [Wong & al. 1991]). Still these proposals can be found to be too constraining, because of the (often implicit) assumption that the certainty degrees of any two pieces of knowledge are comparable. But there might be circumstances under which an expert does not want to compare them, either because it does not make sense, or because he does not have enough information. To overcome this limitation, this paper suggests the use of a partially ordered set of grades to express levels of certainty. A noticeable benefit resulting from such a generalization is the possibility of representing knowledge coming from different sources in a single frame, by means of different scales of grades. This is important because generally two experts do not attribute the same meaning to the same grade.

A major concern when dealing with uncertain knowledge is the way certainty degrees are propagated and combined during the reasoning process. Numerical approaches are often based on calculus systems that may be coupled with deduction systems (e.g., production systems [Shortliffe & Buchanan 1975], belief networks [Pearl 1988] [Shenoy & Shafer 1988]). The semantical characterization of such hybrid systems is not always an easy task. Several formalisms have considered the assignment of certainty degrees to logical formulas (e.g., [Nilsson 1986], [Dubois & al. 1987]). This is more satisfactory from the semantical point of view, but such systems still have two separate components: one for characterizing the logical apparatus and the other for describing the properties of the numerical calculus. Another possibility is to introduce the certainty degree calculus in the logic itself, as in [Halpern & Rabin 1987]. This is the approach followed in this paper. We propose a multimodal framework in which grades correspond to lower bounds of certainty degrees, and are represented by modal operators. It can be considered as an extension of previous work [Froidevaux & Grossetête 1990], [Chatalic & Froidevaux 91], but this new formalism has an increased expressive power and its semantics is better suited to its syntax. Our choice of a modal framework is motivated by our ultimate goal, which is to have the ability to represent various kinds of imperfect knowledge simultaneously. In particular we are interested in representing uncertain and/or incomplete knowledge, which leads to nonmonotonic reasoning techniques. For that purpose, the work of [Siegel & Schwind 1991] seems quite promising.

Even though symbolic grades do not have to satisfy the laws of numerical settings, they have to behave according to some knowledge representation principles. For instance, when uncertain facts are used in some reasoning, we cannot expect the deduced facts to be more certain than each of the facts used in this deduction (*principle 1*). Another expected property is that, as soon as we consider lower bounds of certainty degrees, if a statement can be obtained in several ways, with different grades, its degree of certainty should be at least as high as each of these grades (*principle 2*). These two basic principles also may be found in the framework of possibilistic logic [Dubois & al. 1987]. With partially ordered values, this leads us naturally to consider greatest lower bounds and least upper bounds of grades. As a consequence, the set of all possible grades will be structured in a lattice. Note that several approaches using partially ordered sets of grades have been suggested for the treatment of uncertain and/or incomplete knowledge such as [Rasiowa 1987], [Ginsberg 1988], [Fitting 1991] or [Subrahmanian 1988]. However all these proposals tackle this problem with multivalued logic frameworks, which is not the case in our approach.

The remainder of the paper is organized as follows. In section 2 we present the language of multimodal graded logic, propose an axiomatic system and indicate some of its properties. Section 3 introduces the semantics in terms of graded interpretations. The proposed axiomatization is proved to be sound and complete with respect to this semantics. Section 4 presents a comparison of this work with some related approaches. Finally, we conclude with some directions for further research.

## 2. THE LANGUAGE

In this framework, grades are attached to formulas. We propose to express the knowledge in a multimodal language, in which each grade corresponds to some modal operator.

### 2.1 EXPRESSING GRADES

The prefered way of expressing qualitative uncertainty generally consists in attaching symbolic grades of certainty to pieces of information and in specifying how these grades relate each other. Therefore, to express our initial knowledge we start from an finite set of grades $\Gamma_0 = \{\alpha_1, \ldots, \alpha_n\}$, partially ordered by a relation $\preccurlyeq$. We suppose that $\Gamma_0$ contains a universal upper bound T (i.e. such that $\forall\alpha\in\Gamma_0, \alpha\preccurlyeq T$), which is used to characterize knowledge known as certain.

During reasoning steps, grades attached to pieces of knowledge involved in deductions are combined together, in order to characterize the grades attached to the conclusions. According to our basic principles, this leads to consider upper and lower bounds of grades. This is formalized by means of two binary operators $\wedge$ and $\vee$, called respectively *meet* and *join*. The set $\Gamma_0(\wedge, \vee)$ of expressions over $\Gamma_0$ is then constructed recursively as the smallest set satisfying:

- $\Gamma_0 \subset \Gamma_0(\wedge, \vee)$
- $\forall\alpha,\beta\in \Gamma_0(\wedge, \vee)$, $\alpha\vee\beta\in \Gamma_0(\wedge, \vee)$ and $\alpha\wedge\beta\in\Gamma_0(\wedge, \vee)$

The partial order relation $\preccurlyeq$ on $\Gamma_0$ may be extended to a new relation $\preccurlyeq$ over $\Gamma_0(\wedge, \vee)$, defined as the smallest relation verifying:

$\forall\alpha, \beta, \gamma \in \Gamma_0(\wedge, \vee)$:

- if $\alpha, \beta \in \Gamma_0$ and $\alpha \preccurlyeq \beta$ then $\alpha \preccurlyeq \beta$ (I-1)
- if $\alpha \preccurlyeq \gamma$ and $\beta \preccurlyeq \gamma$ then $\alpha\vee\beta \preccurlyeq \gamma$ (I-2)
- if $\alpha \preccurlyeq \gamma$ or $\beta \preccurlyeq \gamma$ then $\alpha\wedge\beta \preccurlyeq \gamma$ (I-3)
- if $\alpha \preccurlyeq \beta$ and $\alpha \preccurlyeq \gamma$ then $\alpha \preccurlyeq \beta\wedge\gamma$ (I-4)
- if $\alpha \preccurlyeq \beta$ or $\alpha \preccurlyeq \gamma$ then $\alpha \preccurlyeq \beta\vee\gamma$ (I-5)
- $\alpha\wedge(\beta\vee\gamma) \preccurlyeq (\alpha\wedge\beta)\vee(\alpha\wedge\gamma)$ (I-6)

We define $\alpha \equiv \beta$ in $\Gamma_0(\wedge, \vee)$ to mean that $\alpha\preccurlyeq\beta$ and $\beta\preccurlyeq\alpha$. The relation $\equiv$ is clearly an equivalence relation on $\Gamma_0(\wedge, \vee)$. The quotient set $\Gamma_0(\wedge, \vee)/_\equiv$ can be shown to be a distributive lattice, in which the expressions $\alpha\vee\beta$ and $\alpha\wedge\beta$ denote respectively the lowest upper bound (*lub*) and the greatest lower bound (*glb*) of $\alpha$ and $\beta$. The initial set $\Gamma_0$ is called the set of ***generators*** of $\Gamma_0(\wedge, \vee)/_\equiv$ and $(\Gamma_0(\wedge, \vee)/\equiv, \wedge, \vee, \preccurlyeq)$ is called ***the distributive lattice generated by*** $(\Gamma_0, \preccurlyeq)$. For the sake of simplicity $\Gamma_0(\wedge, \vee)/_\equiv$ and $\Gamma_0(\wedge, \vee)$ will be denoted respectively by $\Gamma$ and $\Gamma^*$. We will also use $\Gamma_0$ to denote the subset of $\Gamma$ corresponding to classes of elements of $(\Gamma_0, \preccurlyeq)$.

**Example:**

Let us suppose that our initial knowledge is expressed using grades of the $\Gamma_0 = \{\alpha, \beta, \gamma, \delta, T\}$ such that $\gamma<\alpha$, $\gamma<\delta$ and $\beta<\delta$. The extended partial order on $\Gamma$ is summarized on the following diagram:

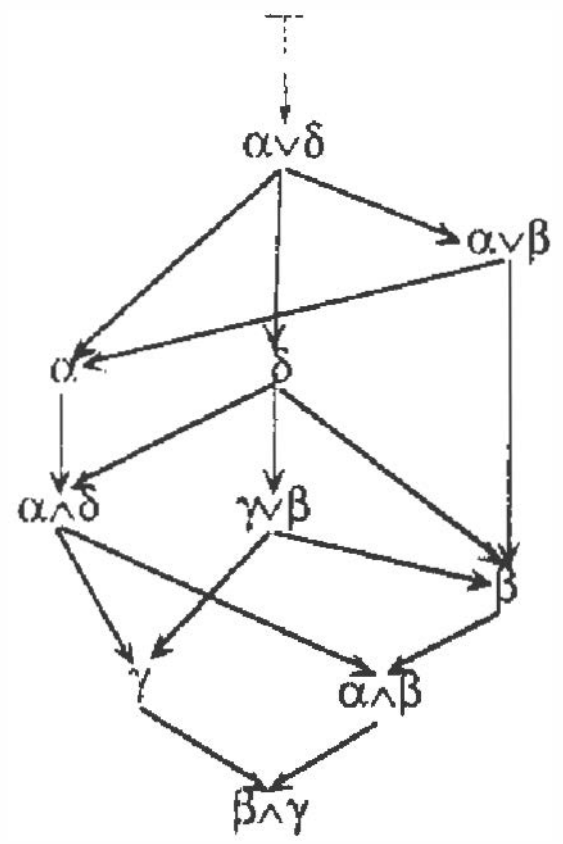

This lattice generated from $(\Gamma_0, \preccurlyeq)$ contains exactly all possible grades that may appear in deduced facts. The reason why we choose to consider a lattice generated by a partially ordered set is that glb and lub of any two non comparable grades of $\Gamma$ are necessarily distinct from elements of $\Gamma_0$.

For instance, if two pieces of knowledge graded respectively by $\alpha$ and $\delta$ are used to deduce a fact with grade $\alpha\wedge\delta$ (according to principle 1), the grade of the conclusion is greater than $\gamma$, which is intuitively satisfactory since the deduction only uses grades strictly greater than $\gamma$.

## 2.2 GRADED FORMULAS

Let $P = \{p_0, p_1, ..., p_n\}$ be a finite set of atomic propositions. As usual $\neg, \wedge, \vee, \rightarrow, \leftrightarrow$ will denote the boolean connectives. Let $(\Gamma, \wedge, \vee, \leqslant)$ be the distributive lattice[1] generated by some finite set of partially ordered grades $(\Gamma_0, \leqslant)$, and $\Gamma*$ the set of expressions on $\Gamma$.

For every expression $\alpha\in\Gamma*$, $[\alpha]$ denotes a modal operator (also called parameterized modal operator). Intuitively, $[\alpha]p$ corresponds to a formula p which is known with at least grade $\alpha$. The symbol $\alpha$ is thus considered as a *lower bound* of the certainty degree of p.

**DEFINITION:** The multi-modal graded **language** $L_{P,\Gamma}$ induced by $P$ and $\Gamma$ is the least set satisfying the following conditions:

- • $P \subset L_{P,\Gamma}$
- •• $\forall$ p, q$\in L_{P,\Gamma}$, $\neg p$, $p\vee q$, $p\wedge q$, $p\rightarrow q$, $p\leftrightarrow q$ are formulas of $L_{P,\Gamma}$
- ••• $\forall p\in L_{P,\Gamma}$, $\forall\alpha\in\Gamma^*$, $[\alpha]p\in L_{P,\Gamma}$

In the following, each element of $\Gamma$ will be denoted by any expression of its equivalence class in $\Gamma*$. We use the symbols $\alpha, \beta, \gamma, ...$ to denote syntactic variables ranging over $\Gamma$.

## 2.3 AXIOM SYSTEM

We now consider a syntactical characterization of our basic principles. It is worth noticing that such an axiomatization corresponds to two embedded calculus: one characterizing the modal propositional calculus and the other one characterizing the distributive lattice structure of the set of grades.

We present an axiom sytem, denoted by $\Sigma_\Gamma$, which is based on the modal system K.

**Axiom schemes:**

(C) Classical axioms schemes

(K) $[\alpha](A \rightarrow B) \rightarrow ([\alpha]A \rightarrow [\alpha]B)$

(D-) $\neg[T]$false.

(A1) $([\alpha]A \wedge [\beta]A) \rightarrow [\alpha\vee\beta]A$

(A2) $([\alpha]p_0 \vee [\beta]p_0) \rightarrow [\alpha\wedge\beta]p_0$

(A3) $[\alpha\vee\beta]p_0 \rightarrow ([\alpha]p_0 \wedge [\beta]p_0)$

(A4) $[(\alpha\wedge\beta)\vee(\alpha\wedge\gamma)]p_0 \rightarrow [\alpha\wedge(\beta\vee\gamma)]p_0$

(A5) axioms of the form $[\alpha]p_0 \rightarrow [\beta]p_0$ $\forall\alpha, \beta\in\Gamma_0$ such that $\beta < \alpha$.

**Inference rules:**

$(R_0)\ \dfrac{\vdash A \quad \vdash A \rightarrow B}{\vdash B}$ (modus ponens rule)

$(R_1)\ \dfrac{\vdash A}{\vdash [T]A}$ (necessitation rule)

$(R_2)\ \dfrac{\vdash [\beta]p_0 \rightarrow [\alpha]p_0 \text{ and } \vdash [\gamma]p_0 \rightarrow [\alpha]p_0}{\vdash [\beta\wedge\gamma]p_0 \rightarrow [\alpha]p_0}$ (glb rule)

$(R_3)\ \dfrac{\vdash [\beta]p_0 \rightarrow [\alpha]p_0}{\vdash [\beta]A \rightarrow [\alpha]A}$ (generalization rule)

Note that in this system, the symbols $\alpha$, $\beta$ and $\gamma$ refer to any expression of $\Gamma*$ (except for $A_5$). Similarly, symbols A and B refer to any formulas, but this is not the case for $p_0$, which corresponds to some specific atomic proposition in $P$, as explained below. In the following, a theorem p of $\Sigma_\Gamma$ will be denoted by $\vdash_\Gamma p$ or more simply by $\vdash p$ if there is no ambiguity.

For a correct understanding of $\Sigma_\Gamma$ it is essential to distinguish the part characterizing the structure of the set of modal operators, from the rest of the system. In $\Sigma_\Gamma$, we express the fact that a grade $\alpha$ is greater than a grade $\beta$, by a theorem of the form $[\alpha]p_0 \rightarrow [\beta]p_0$. It should be stressed that the symbol $p_0$ corresponds here to some ***specific*** atomic proposition in $P$, that can only be used for characterizing the partial order on $\Gamma$. The distributive lattice structure of $\Gamma$ is then expressed by means of axiom schemes $A_1$ to $A_5$ and the inference rule $R_2$. Axioms of the form $A_5$, express the partial order on the initial set of grades $\Gamma_0$ and correspond to the relation (1-1) introduced in section 2.1 (strict ordering is sufficient since $p \rightarrow p$ is a theorem of classical logic). $A_1$, $A_2$, $A_3$ and $R_2$ characterize properties of the meet and join operations on $\Gamma$. They should be related to the relations (1-2 to 1-5) introduced in section 2.1. $A_4$ confers the distributivity property and corresponds to (1-6). The need for this property will be justified shortly.

Notice that the introduction of a special atomic formula $p_0$ for characterizing the partial order on $\Gamma$ is essential, since the rule $R_2$ would not be correct if stated for any formula A (see section 3). It is also essential to consider $p_0$ as a ***reserved*** atomic formula, which in no case may be used for expressing some particular knowledge in a given theory.

A special attention must also be paid to the particular axiom scheme $A_1$, which is actually stronger than necessary. In fact, instead of $A_1$, an inference rule like:

$(R_{2bis})\ \dfrac{\vdash [\alpha]p_0 \rightarrow [\beta]p_0 \text{ and } \vdash [\alpha]p_0 \rightarrow [\gamma]p_0}{\vdash [\alpha]p_0 \rightarrow [\beta\vee\gamma]p_0}$

could have been sufficient to characterize the distributive lattice structure of $\Gamma$. But the contribution of $A_1$ is more important. It also expresses our second basic principle which states that, if there are several ways to deduce a

[1] we use the same symbols $\wedge$ and $\vee$ to denote meet and join operations on $\Gamma$ and the logical connectors.

given proposition with different grades, the best known grade for this proposition should be at least as high as any of those grades. This is the reason why the $A_1$ axiom scheme refers to any formula A and not merely to the specific atomic formula $p_0$. More generally, in the case where there are several ways of deriving a same proposition p with different grades, $A_1$ will be used to obtain the ***best*** possible grade, i.e. the ***greatest lower bound*** of the certainty degree of p.

A first result concerning the correspondence between the partial order $\leqslant$ on $\Gamma$, and its counterpart in $\Sigma_\Gamma$ is:

**THEOREM 1**: *$\forall \alpha, \beta \in \Gamma$, if $\alpha \leqslant \beta$ then $\vdash [\beta]p_0 \rightarrow [\alpha]p_0$*

The proof proceeds by induction on the complexity of $\alpha$ and $\beta$.

From this, the generalization rule $R_3$ merely expresses that modal operators correspond to lower bounds of certainty degrees. This implies that if $\alpha \leqslant \beta$, the theorem $[\beta]p \rightarrow [\alpha]p$ holds for any formula p. This may be related to the ***weakening rule*** introduced in [Chatalic & Froidevaux 1991]:

$$\frac{\vdash [\alpha]p}{\vdash [\beta]p \quad \forall \beta \leqslant \alpha}$$

The system $\Sigma_\Gamma$ also includes the axiom D.. . This axiom rejects theories from which the contradiction may be derived with full certainty. Notice that $\Sigma_\Gamma$ does not prevent us from considering partially inconsistent theories, where the contradiction could be derived with some grade $\alpha < T$. This is satisfactory from the knowledge representation point of view, since with uncertain knowledge we often have to deal with conflicting pieces of information [Dubois & al. 1991]. As a consequence, [T] must be considered as the necessity operator of a system D, while for $\alpha < T$, $[\alpha]$ is the necessity operator of a system K [Chellas 1980].

## 2.4 SOME PROPERTIES OF $\Sigma_\Gamma$

• As a consequence of theorem 1, usual properties of distributive lattices may be obtained as theorems or derived inference rules of $\Sigma_\Gamma$. For instance we have:

$\vdash [\alpha]p_0 \rightarrow [\alpha \wedge \beta]p_0$

$\vdash [\alpha \vee \beta]p_0 \rightarrow [\alpha]p_0$

$\vdash [\alpha \wedge \beta]p_0 \rightarrow [\beta \wedge \alpha]p_0$ and $\vdash [\alpha \vee \beta]p_0 \rightarrow [\beta \vee \alpha]p_0$

$$\frac{\vdash ([\beta]p_0 \rightarrow [\gamma]p_0)}{\vdash [\alpha \wedge \beta]p_0 \rightarrow [\alpha \wedge \gamma]p_0}$$

•• Another interesting property is that we obtain as a derived axiom scheme (using K, $A_2$ and modus ponens):

$$\vdash ([\alpha]A \wedge [\beta](A \rightarrow B)) \rightarrow [\alpha \wedge \beta]B \qquad (A_G)$$

This axiom scheme clearly expresses our first basic principle. It allows for an easily formalization of deductions in theories with uncertain knowledge and is related to the ***graded modus ponens*** rule introduced in [Chatalic & Froidevaux 1991]:

$$\frac{\vdash [\alpha]p \quad \vdash [\beta](p \rightarrow q)}{\vdash [\alpha \wedge \beta]q} \qquad \text{(g.m.p.)}$$

Given a set S of formulas, we say that a formula p is *deducible* or *derivable* from S (written $S \vdash p$) in $\Sigma_\Gamma$ if and only if we can find a finite subset $\{q_1, \ldots, q_n\} \subseteq S$ such that $(q_1 \wedge \ldots \wedge q_n) \rightarrow p$ is a theorem of $\Sigma_\Gamma$. For simplicity, the conjunction of formulas of S is denoted by $S_\wedge$ in the following.

Let us consider a first example mixing both principles.

**Example 1:** Let us suppose that we have some reasons to believe that the weather will be cold and rainy tomorrow and that both aspects may have an influence on our health state. This is expressed as the set

$S = \{[\alpha]cold, [\beta]rain, [\gamma](cold \rightarrow ill), [\delta](rain \rightarrow ill)\}$

Let us assume we cannot compare $\alpha$, $\gamma$, $\beta$ and $\delta$. Then using $(A_G)$ we have:

1 - $\vdash ([\alpha]cold \wedge [\gamma](cold \rightarrow ill)) \rightarrow [\alpha \wedge \gamma]ill.$ $(A_G)$

2 - $\vdash ([\beta]rain \wedge [\delta](rain \rightarrow ill)) \rightarrow [\beta \wedge \delta]ill.$ $(A_G)$

and thus using $(A_1)$ and properties of classical logic:

3 - $\vdash S_\wedge \rightarrow [(\alpha \wedge \gamma) \vee (\beta \wedge \delta)]ill$ i.e

$S_\wedge \vdash [(\alpha \wedge \gamma) \vee (\beta \wedge \delta)]ill$ .

wich is the greatest lower bound of the certainty degree we can derive for the proposition *ill*.

The second example justifies the need for the distributivity property.

**Example 2:** Let us suppose now that we have two different reasons to believe that the weather will be cold with different degrees $\alpha$ and $\beta$. Thus we have

$S = \{ [\alpha]cold, [\beta]cold, [\gamma](cold \rightarrow ill) \}$

We still assume that $\alpha$, $\gamma$, $\beta$ are not comparable. We consider the two following graded deductions:

a) 1 - $\vdash ([\alpha]cold \wedge [\gamma](cold \rightarrow ill)) \rightarrow [\alpha \wedge \gamma]ill$ $(A_G)$

2 - $\vdash ([\beta]cold \wedge [\gamma](cold \rightarrow ill)) \rightarrow [\beta \wedge \gamma]ill$ $(A_G)$

hence $S_\wedge \vdash [(\alpha \wedge \gamma) \vee (\beta \wedge \gamma)]ill$ (A1 and m.p.)

b) 1 - $\vdash ([\alpha]cold \wedge [\beta]cold) \rightarrow [\alpha \vee \beta]cold$ (A1)

2 - $\vdash ([\alpha \vee \beta]cold \wedge [\gamma](cold \rightarrow ill)) \rightarrow$

$[(\alpha \vee \beta) \wedge \gamma]ill$ $(A_G)$

hence $S_\wedge \vdash [(\alpha \vee \beta) \wedge \gamma]ill$

But since the lattice is distributive, we have: $(\alpha \wedge \gamma) \vee (\beta \wedge \gamma) = (\alpha \vee \beta) \wedge \gamma$. More generally, the distributivity property makes it possible to apply the inference rules in any order.

The third example motivates the use of glb for non comparables grades.

**Example 3:** In this example John must go to the theater and is affraid of possible traffic jams wich might probably cause him to arrive late. Moreover he has a lot of work and thus a few chances to finish his work early. But if it is possible, he might go to the restaurant prior to the theater. This may be formalized by:

$S = \{ [\alpha]traffic_jams,\ [\delta]finish_work_early,$
$\quad [\beta](traffic_jams \rightarrow late),$
$\quad [\gamma](finish_early \rightarrow restaurant)\ \}.$

This time we assume that $\gamma \preccurlyeq \alpha$ and $\delta \preccurlyeq \beta$ and nothing else. We obtain:

1 - $\vdash\ ([\alpha]traffic_jams \wedge [\beta]([\alpha]traffic_jams \rightarrow late)) \rightarrow [\alpha\wedge\beta]late \qquad (A_G)$

2 - $\vdash\ ([\delta]finish_early,\ [\gamma](finish_early \rightarrow restaurant)) \rightarrow [\delta\wedge\gamma]restaurant \qquad (A_G)$

hence $S_\wedge \vdash [\alpha\wedge\beta]late \wedge [\delta\wedge\gamma]restaurant$ (m.p.)
Notice that $\alpha$ and $\beta$ (respectively $\gamma$ and $\delta$) are not comparable, but that we are still able to compare the best lower bounds for *late* and *restaurant*. And since $\delta\wedge\gamma \preccurlyeq \alpha\wedge\beta$ we have more confidence into the fact that John will arrive late at the theater than into the fact that he will go to the restaurant.

**Example 4:** The last example shows how the formalism we present here, is more expressive than those in the previous approaches. In this example, some agent John expresses his degree of confidence $\alpha$ into the point of view of agent Mike (degrees of certainty $\beta$ and $\gamma$). John is almost certain that if Mike is certain that Tom will not come tonight, then Mike thinks that it is highly likely that Mary will not come. Moreover, John is certain (degree of certainty $\alpha'$) that Mike is certain that Tom will not come tonight. We assume that $\alpha \preccurlyeq \alpha'$.
We get the set of formulas:
$S = \{ [\alpha]\ ( [\beta] \neg\ tom_coming \rightarrow [\gamma] \neg mary_coming),$
$\quad [\alpha'] [\beta] \neg\ tom_coming\ \}.$

By theorem 1, we get:
$\vdash [\alpha'] [\beta] \neg\ tom_coming \rightarrow [\alpha] [\beta] \neg\ tom_coming$

Then using (K) we have: $\vdash S_\wedge \rightarrow$
$([\alpha] [\beta] \neg\ tom_coming \rightarrow [\alpha] [\gamma] \neg mary_coming)$

Therefore: $\vdash S_\wedge \rightarrow [\alpha] [\gamma] \neg mary_coming.$

In conclusion, John is almost certain that Mike thinks that it is highly likely that Mary will not come.

## 3 SEMANTICS

We define the meaning of formulas by a possible worlds semantics [Chellas 1980] involving families of accessibility relations. In fact, with each grade $\alpha\in\Gamma$, we associate an accessibility relation $R_\alpha$. Our intuition is that the higher the grade associated with a formula, the more constraining the corresponding accessibility relation should be. We express this idea by the fact that if $\alpha \preccurlyeq \beta$ then $R_\alpha \subseteq R_\beta$. In such a case, if, from a given possible world w, there are more possible worlds accessible through $R_\beta$ than through $R_\alpha$, it will be more difficult to satisfy $[\beta]p$ than $[\alpha]p$ in w.

**DEFINITION:** A $\Gamma$-interpretation is defined as a triple $I = \langle W, (R_\alpha)_{\alpha\in\Gamma}, s \rangle$ where:

- W is a set of worlds
- $(R_\alpha)_{\alpha\in\Gamma}$ is a family of accessibility relations verifying:
  - i) for each pair $(\alpha, \beta)\in\Gamma$: if $\alpha \preccurlyeq \beta$ then $R_\alpha \subseteq R_\beta$
  - ii) $\forall\alpha, \beta \in \Gamma$, $R_{\alpha\vee\beta} = R_\alpha \cup R_\beta$
  - iii) $\forall\alpha, \beta \in \Gamma$, $R_{\alpha\wedge\beta} = R_\alpha \otimes R_\beta$
  - iv) $R_\top$ is serial (i.e., $\forall w\in W$, $\exists w'\in W$ such that $R_\top(w,w')$ )
- $s: P\times W \rightarrow \{true, false\}$ is a classical truth value assignment

The symbol $\otimes$ denotes glb operation on the set of accessibility relations (with respect to set inclusion) and $\cup$ denotes classical set union. Note that $R_\alpha \cup R_\beta$ corresponds to the lub of $R_\alpha$ and $R_\beta$. As a consequence $((R_\alpha)_{\alpha\in\Gamma}, \otimes, \cup, \subseteq)$ has a lattice structure.

**DEFINITION:** Let $I = \langle W, (R_\alpha)_{\alpha\in\Gamma}, s \rangle$ be a $\Gamma$-interpretation. A formula f of $L_{P,\Gamma}$ is said to be **true at a world** w of I (written $I,w \models f$) iff:

- $I,w \models f$ iff $s(f,w) = true$, for $f \in P$
- $I,w \models f\vee g$ iff $I,w \models f$ or $I,w \models g$, for $f, g \in L_{P,\Gamma}$
- $I,w \models f\wedge g$ iff $I,w \models f$ and $I,w \models g$, for $f, g \in L_{P,\Gamma}$
- $I,w \models \neg f$ iff not $I,w \models f$ (that will be denoted by $I,w \not\models f$), for $f \in L_{P,\Gamma}$
- $I,w \models [\alpha]f$ iff $\forall w'\in W$, $R_\alpha(w,w')$ implies $I,w' \models f$, for $f \in L_{P,\Gamma}$

In the following, when there is no ambiguity, the interpretation I will not be mentioned any more and we shall merely write $w \models f$ instead of $I,w \models f$.

**DEFINITIONS:**

- A formula f is **consistent** or **satisfiable** iff there exist a $\Gamma$-interpretation I and a world w such that $I,w \models f$.
- A $\Gamma$-interpretation $I = \langle W, (R_\alpha)_{\alpha\in\Gamma}, s \rangle$ is a **model** of a formula f (written $I \models f$) iff: $\forall w\in W$, $I, w \models f$. We also say that f is **valid in I**.

- A formula f is **valid** (written $\models$ f) iff every Γ–interpretation is a **model** of f.

The previous definitions may be extended to sets of formulas.

Note that the relation $R$ has to be serial. This is required by the presence of the axiom A6 in $\Sigma_\Gamma$.

**THEOREM 2:**
*$\forall \alpha, \beta \in \Gamma$, if $\alpha \preccurlyeq \beta$ then $\models [\beta]p_0 \rightarrow [\alpha]p_0$*
*Conversely if $\models [\beta]p_0 \rightarrow [\alpha]p_0$ then $R_\alpha \subseteq R_\beta$*

SKETCH OF PROOF:

1) $\forall \alpha,\beta \in \Gamma$, $\alpha \preccurlyeq \beta$ implies $\models |\beta|p_0 \rightarrow |\alpha|p_0$.
   Let $\alpha,\beta \in \Gamma$ be such that $\alpha \preccurlyeq \beta$, and let $I = \langle W, (R_\alpha)_{\alpha \in \Gamma}, s \rangle$ be a Γ-interpretation.
   Let w, w'∈W be such that $w \models |\beta|p_0$ and $R_\alpha(w,w')$.
   Since $\alpha \preccurlyeq \beta$, $R_\alpha \subseteq R_\beta$, thus we must also have $R_\beta(w,w')$. From $w \models |\beta|p_0$ we obtain $w' \models p_0$. Thus $\forall w' \in W$, $R_\alpha(w,w')$ implies $w' \models p_0$, i.e., $w \models |\alpha|p_0$.
   Hence $w \models |\beta|p_0 \rightarrow |\alpha|p_0$, for any w∈W, and any Γ-interpretation I.

2) $\forall \alpha,\beta \in \Gamma$, $\models |\beta|p_0 \rightarrow |\alpha|p_0$ implies $R_\alpha \subseteq R_\beta$.
   If $\beta = \top$ it is clear that $\alpha \preccurlyeq \top$. Otherwise, the proof proceeds by exhibiting some particular graded interpretation in which $|\beta|p_0 \rightarrow |\alpha|p_0$ is not valid if not $R_\alpha \subseteq R_\beta$.

**THEOREM 3 (Soundness):** *If $\vdash$ f then $\models$ f.*

SKETCH OF PROOF: The soundness of $\Sigma_\Gamma$ is proved as usual, by induction on the length of derivations. We first prove that each axiom scheme is valid, and then prove the soundness of each inference rule.

We may check at this stage that rule R2 would not be sound if stated for any proposition A, since in the case where $R_\alpha \subseteq R_{\beta\wedge\gamma}$ does not hold, it is possible to construct a Γ-interpretation in which $|\beta|A \rightarrow |\alpha|A$ and $|\gamma|A \rightarrow |\alpha|A$ would hold but not $|\beta\wedge\gamma|A \rightarrow |\alpha|A$ (this happens for instance if in some world we have $\neg|\beta|A$, $\neg|\gamma|A$, $\neg|\alpha|A$ but $|\beta\wedge\gamma|A$)

**THEOREM 4 (Completeness):** *If $\models$ f then $\vdash$ f.*

SKETCH OF PROOF: The completeness result is established by using the Henkin method. The proof proceeds by exhibiting a specific "canonical" interpretation $I_c$ defined on the set of maximally consistent sets [Chellas 1980]. The main point in the proof is to show that this canonical interpretation is a graded interpretation. First we show that in this canonical interpretation we have $\forall \alpha,\beta \in \Gamma$, if $\alpha \preccurlyeq \beta$ then $R_\alpha \subseteq R_\beta$. Then the other properties follow naturally. We refer to [Chatalic & Froidevaux 1992] for details of the proofs.

We are now investigating the links between multimodal graded logic and other close approaches.

# 4 RELATED WORK

Another lattice based logical formalism for handling uncertain knowledge, called *graded logic*, has been introduced in [Chatalic & Froidevaux 1991]. Graded logic deals with formulas of the form (f α), where f is a classical formula and α is a grade. Such an approach falls into the more general setting of *labelled deduction systems* presented in [Gabbay 1991]. The latter allows for a description of the inference apparatus of graded logic (including graded modus ponens), but is merely a formal tool for performing deduction.

In contrast, multimodal logic enjoys all the features of modal logic. It allows for implications between graded formulas like $|\alpha|$ p → $|\beta|$ q, and for embedding of degrees of certainty like $|\alpha_1||\alpha_2|$, so that its expressive power is increased. This latter possiblity is particularly interesting for modelling information coming from several sources. For example, to represent the knowledge of two distinct agents $A_1$ and $A_2$, we will consider in Γ two families of parameters $\Gamma_1$ and $\Gamma_2$, since the agents do not necessarily refer to the same scale. In this context, formula $|\alpha_1||\alpha_2|$ f will mean that agent $A_1$ has a level of confidence $\alpha_1$ in the fact that agent $A_2$ has a level of confidence $\alpha_2$ in fact f. Eventually, as the new language considered is a multimodal one, the Kripke-semantics used is well- suited to the syntax.

Our treatment of uncertainty is closely related to possibilistic logic, a formalism introduced by Dubois and al. [Dubois & al. 1987] as a numerical logic of uncertainty. Possibilistic logic handles uncertainty by means of two dual measures : necessity and possibility measures, both mapping the set of sentences into the interval [0,1]. The semantics of possibilistic logic is defined by using fuzzy sets of classical interpretations. Our approach adopts and generalizes the principles that govern necessity measures, by considering that the grades are not necessarily numerical values and no longer constitute a totally ordered set. Formal links between both formalisms have been established in [Froidevaux & Grossetête 1990]. Such a formal correspondence is enabled by the correspondence between possibility theory and qualitative possibility relations proved in [Dubois 86].

The theory of qualitative necessity measures is closely related to some other formalisms. First, the notion of epistemic entrenchment [Gärdenfors 1988], used for restoring consistency in a knowledge base after updates, has been shown to be linked to possibilistic logic in [Dubois & Prade 1991]. Since [Gärdenfors & Makinson 1988] uses a partial order, the link with multimodal graded logic is stronger. Secondly, [Farinas and Herzig 1991] throws some light on the relations between qualitative possibility theory and modal logics. One point concerns the case where the total order is considered as a conditional connective. Another point deals with the presentation of a multimodal logic in the same way as our approach. In [Farinas and Herzig 1991] the axiomatization about the interaction between modal operators expresses the fact that the set of parameters is totally ordered. Therefore, our formalism strictly extends Farinas and Herzig's approach since the notion of lattice is more general.

Another approach to qualitative necessity measures [Gärdenfors 1991] concerns the definition of nonmonotonic inference relations. For that purpose, orderings between certainty degrees on formulas (called *expectation orderings*) are used. Recent work by Farinas et al. [Farinas and al. 1992] extends the definition of nonmonotonicity to the case of incomplete expectation orderings. Because this approach is also based on a partial order, multimodal graded logic also could be used for defining nonmonotonic inference relations.

Multimodal graded logic, as presented here, could be described in the general setting of [Ohlbach & Herzig 1991]. While their proposal is more general, our approach allows one to describe the structure of the parameter set in the logical syntax. Moreover, we do not assume the partial order to be explicitly known in the whole lattice (it can be deduced from the axiomatic system). We do not have to interpret any chain of modal operators in terms of another modal operator, unlike [Ohlbach & Herzig 1991] who also want to be able to interpret parameters as probability values. Therefore, standard relational Kripke semantics is sufficient for our approach.

Let us mention another modal approach to uncertainty that attempts to capture qualitative reasoning about probability through reasoning about likelihood. In [Halpern & Rabin 1987], degrees of likelihood are formalized by chains of modal operators, built by means of three modal operators. The principles underlying the likelihood logic are therefore deeply different from ours.

The expression graded modal logic refers to another formalism in the recent work of [Van der Hoek 1992]. Beyond the pure analogy between the terms, the formalism proposed in [Van der Hoek 1992] bears some similarities with ours as far as it is also motivated by the representation of uncertain knowledge by means of modal operators. However, this formalism is distinguished from ours in a fundamental way, since it aims at counting the number of exceptional situations where some proposition p does not hold. For this, an infinity of necessity operators $L_n$ ($n \in \mathbb{N}$) is introduced, such that $L_n$ p is satisfied by a (possible) world w if and only if there are at most n worlds w' that are reachable from w and that satisfy formula $\neg p$.

## CONCLUSION AND PERSPECTIVES

In this paper we have presented a formal system which uses a partially ordered set of grades to represent degrees of certainty. Uncertain information is expressed by means of parameterized modal operators. A major feature of this system is that both the treatment of certainty degrees and the characterization of the structure of the set of parameters, are embedded in the axiomatization. We have given a possible worlds semantics and have established soundness and completeness results.

As mentioned in the introduction, our ultimate aim is to formalize reasoning with both uncertain and incomplete knowledge (see [Murakami & Aibara 1990], [Froidevaux & Grossetête 1990]). We briefly indicate how this goal may be achieved with the current formalism.

Among the numerous logical formalisms addressing the problem of nonmonotonic reasoning, Siegel's and Schwind's *modal logic of hypotheses* [Siegel & Schwind 1991] seems to be very promising. This formalism uses two necessity modal operators L and L', such that Lp means *p is known* and $\neg L'\neg p$ (also denoted by Hp) means *p is assumed*. The two operators are linked by the axiom scheme $\vdash L\neg p \rightarrow \neg Hp$, the intuitive meaning of which is that knowing $\neg p$ prevents us from assuming p. A general rule with exception, like *Generally birds fly except for penguins*, is translated into the formula: $\forall x\ ((L\ bird(x) \wedge H\neg penguin(x)) \rightarrow L\ flies(x))$.

Roughly speaking, the sets of nonmonotonic theorems (called *extensions*) are obtained by maximizing the sets of hypotheses which may be added to the theory while preserving consistency. In order to model uncertainty, we could introduce a family of modal operators $[\alpha]$ instead of L. Links between modal operators would then be specifed by axiom schemes of the form $\vdash [\alpha]\neg p \rightarrow \neg Hp$. Since the inference mechanism of [Siegel & Schwind 91] is compatible with our approach, we could extend the notion of extension into a notion of graded extension.

Another direction for further research about nonmonotonicity following our approach could be to investigate the relations between incomplete expectation orderings, as introduced in [Farinas and al. 92] and multimodal graded logic.

In conclusion, the work presented in this paper can be considered as a further step towards a better understanding of the links between symbolic and quantitative approaches to uncertainty.

**Acknowledgments**

We thank Luis Farinas del Cerro and Serenella Cerrito for many fruitful discussions and useful advice. This work has been partially supported by DRUMS Esprit

Basic Research Action and by national PRC-GRECO Intelligence Artificielle project.